\documentclass[11pt,a4paper]{article}
\usepackage[hyperref]{acl2018}
\usepackage{times}
\usepackage{latexsym}
\usepackage{times}
\usepackage{amsfonts}
\usepackage{latexsym}
\usepackage{url}
\usepackage{makecell}
\usepackage{hhline}
\usepackage{multirow, rotating}
\usepackage{bbding}
\usepackage{amsfonts, amssymb, amsmath}
\usepackage{algorithm, algorithmic, amsthm}
\usepackage{graphicx}
\usepackage{color}
\usepackage{float}
\usepackage{paralist}

\usepackage{subfigure}
\usepackage{todonotes}

\usepackage{url}

\aclfinalcopy 

\usepackage{eqparbox}

\title{Deep Dyna-Q: Integrating Planning for \\ Task-Completion Dialogue Policy Learning}

\author{Baolin Peng$^{\star}$\quad Xiujun Li$^{\dagger}$\quad Jianfeng Gao$^{\dagger}$\quad Jingjing Liu$^{\dagger}$\quad Kam-Fai Wong$^{\star\ddag}$\quad Shang-Yu Su$^{\S}$  \\
  $^{\dagger}$Microsoft Research, Redmond, WA, USA \\
  $^{\star}$The Chinese University of Hong Kong, Hong Kong \\
  $^{\ddag}$MoE Key Lab of High Confidence Software Technologies, China\\
  $^{\S}$National Taiwan University, Taipei, Taiwan \\
  {\tt \{xiul,jfgao,jingjl\}@microsoft.com} \\
  {\tt \{blpeng,kfwong\}@se.cuhk.edu.hk\quad  shangyusu.tw@gmail.com}
}

\date{}

\begin{document}
\maketitle

\begin{abstract}
Training a task-completion dialogue agent via reinforcement learning (RL) is costly 
because it requires many interactions with real users. One common alternative is to use a user simulator. However, a user simulator usually lacks the language complexity of human interlocutors and the biases in its design may tend to degrade the agent. 
To address these issues, we present Deep Dyna-Q, which to our knowledge is the first deep RL framework that integrates planning for task-completion dialogue policy learning.
We incorporate into the dialogue agent a model of the environment, referred to as the \emph{world model}, to mimic real user response and generate simulated experience. During dialogue policy learning, the world model is constantly updated with real user experience to approach real user behavior, and in turn, the dialogue agent is optimized using both real experience and simulated experience. The effectiveness of our approach is demonstrated on a movie-ticket booking task in both simulated and human-in-the-loop settings\footnote{The source code of this work is available at \url{https://github.com/MiuLab/DDQ}}.

\end{abstract}

\section{Introduction}
\begin{figure*}[ht!]
\centering 
\subfigure[Learning with real users] { \label{fig:learning_typeA} 
\includegraphics[width=0.66\columnwidth]{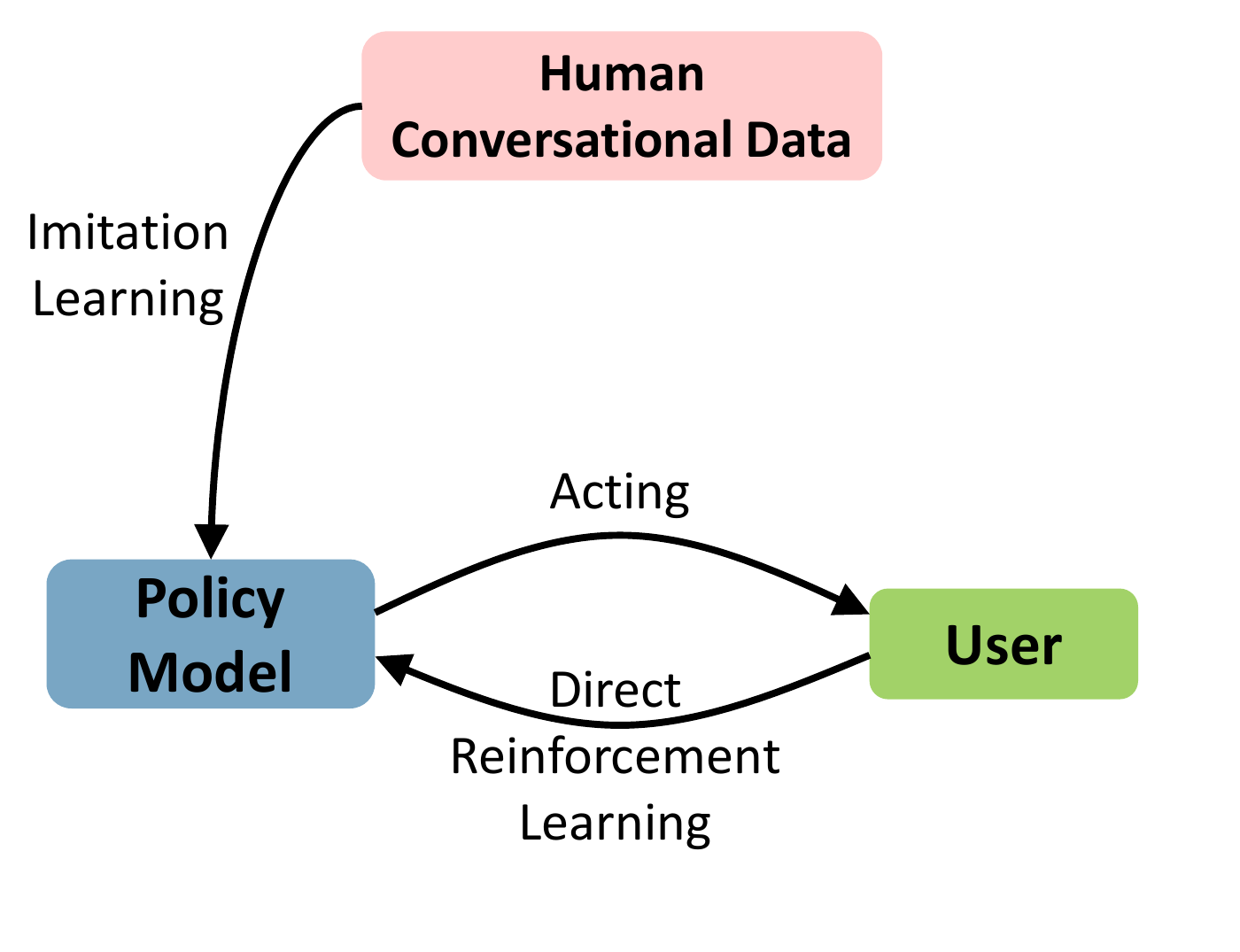}}
\subfigure[Learning with user simulators] { \label{fig:learning_typeB} 
\includegraphics[width=0.66\columnwidth]{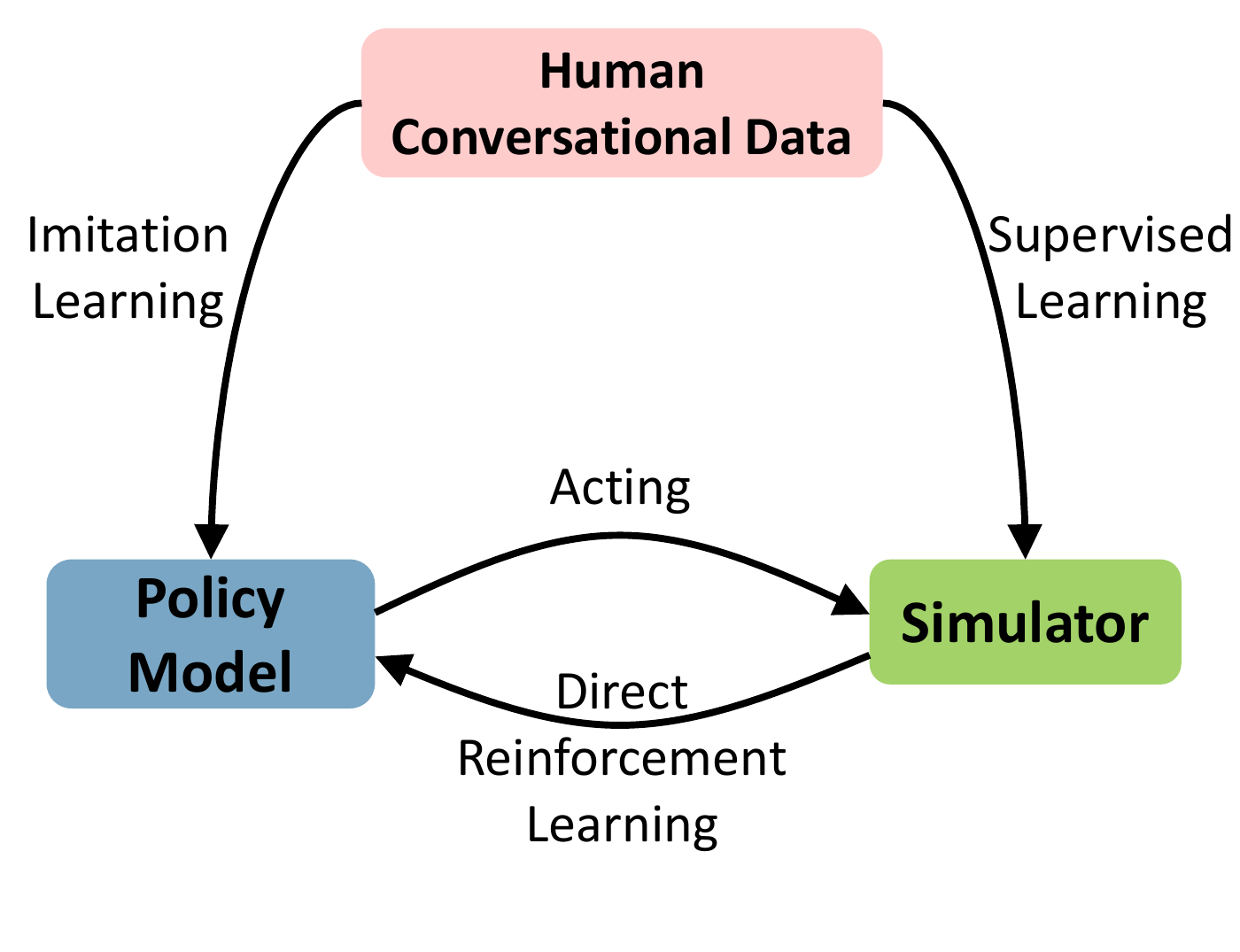}}
\subfigure[Learning with real users via DDQ] { \label{fig:learning_typeC} 
\includegraphics[width=0.66\columnwidth]{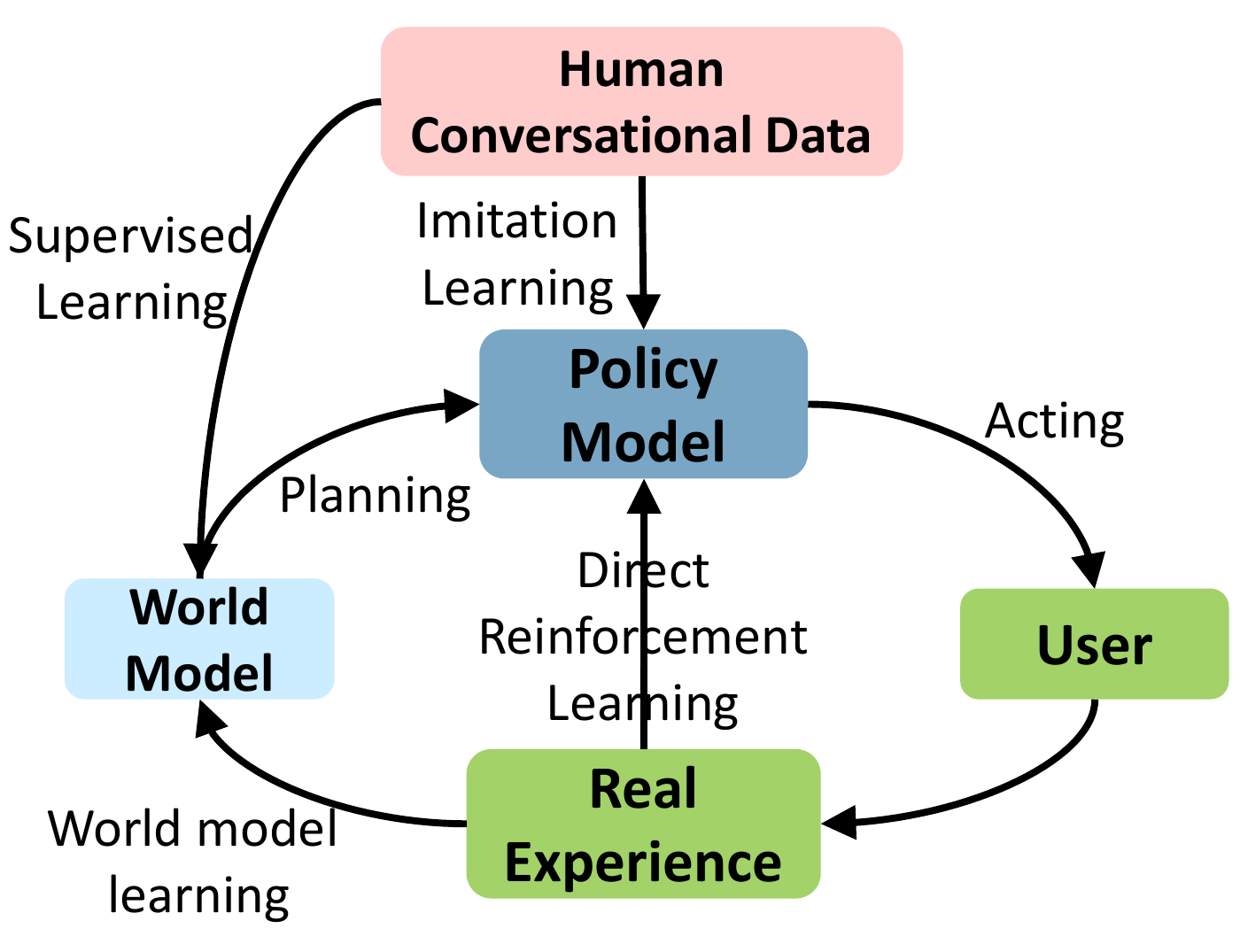}}
\vspace{-2mm}
\caption{Three strategies of learning task-completion dialogue policies via RL.} 
\label{fig:training_types} 
\vspace{-2mm}
\end{figure*}

Learning policies for task-completion dialogue is often formulated as a reinforcement learning (RL) problem~\cite{young2013pomdp,levin1997learning}. However, applying RL to real-world dialogue systems can be challenging, due to the constraint that an RL learner needs an environment to operate in. In the dialogue setting, this requires a dialogue agent to interact with real users and adjust its policy in an online fashion, as illustrated in Figure~\ref{fig:learning_typeA}. Unlike simulation-based games such as Atari games~\cite{mnih2015human} and AlphaGo~\cite{silver2016mastering,silver2017mastering} where RL has made its greatest strides, task-completion dialogue systems may incur significant real-world cost in case of failure. 
Thus, except for very simple tasks~\cite{singh2002optimizing,gavsic2010gaussian,gavsic2011line,pietquin2011sample,li2016dialogue,su2016line}, RL is too expensive to be applied to real users to train dialogue agents from scratch.

One strategy is to convert human-interacting dialogue to a simulation problem (similar to Atari games), by building a user simulator using human conversational data~\cite{schatzmann2007agenda,li2016user}. In this way, the dialogue agent can learn its policy by interacting with the simulator instead of real users (Figure~\ref{fig:learning_typeB}). The simulator, in theory, does not incur any real-world cost and can provide unlimited simulated experience for reinforcement learning. The dialogue agent trained with such a user simulator can then be deployed to real users and further enhanced by only a small number of human interactions. Most of recent studies in this area have adopted this strategy~\cite{su2016continuously,lipton2016efficient,zhao2016towards,williams2017hybrid,dhingra2017towards,li2017end,liu2017iterative,peng2017composite,budzianowski2017sub,peng2017adversarial}.

However, user simulators usually lack the conversational complexity of human interlocutors, and the trained agent is inevitably affected by biases in the design of the simulator. \citet{dhingra2017towards} demonstrated a significant discrepancy in a simulator-trained dialogue agent when evaluated with simulators and with real users. Even more challenging is the fact that there is no universally accepted metric to evaluate a user simulator~\cite{pietquin2013survey}. Thus, it remains controversial whether training task-completion dialogue agent via simulated users is a valid approach.

We propose a new strategy of learning dialogue policy by interacting with real users. Compared to previous works~\cite{singh2002optimizing,li2016dialogue,su2016line,papangelis2012comparative}, our dialogue agent learns in a much more efficient way, using only a small number of real user interactions, which amounts to an affordable cost in many nontrivial dialogue tasks.

Our approach is based on the Dyna-Q framework~\cite{sutton1990integrated} where planning is integrated into policy learning for task-completion dialogue. Specifically, we incorporate a model of the environment, referred to as the \emph{world model}, into the dialogue agent, which simulates the environment and generates simulated user experience. During the dialogue policy learning, real user experience plays two pivotal roles: first, it can be used to improve the world model and make it behave more like real users, via supervised learning; second, it can also be used to directly improve the dialogue policy via RL. The former is referred to as \textit{world model learning}, and the latter \textit{direct reinforcement learning}. Dialogue policy can be improved either using real experience directly (i.e., direct reinforcement learning) or via the world model indirectly (referred to as \textit{planning} or \textit{indirect reinforcement learning}). The interaction between world model learning, direct reinforcement learning and planning is illustrated in Figure~\ref{fig:learning_typeC}, following the Dyna-Q framework~\cite{sutton1990integrated}.

The original papers on Dyna-Q and most its early extensions used tabular methods for both planning and learning ~\cite{singh1992reinforcement,peng1993efficient,moore1993prioritized,kuvayev1996model}. This table-lookup representation limits its application to small problems only. \citet{sutton2012dyna} extends the Dyna architecture to linear function approximation, making it applicable to larger problems. In the dialogue setting, we are dealing with a much larger action-state space. Inspired by~\citet{mnih2015human}, we propose Deep Dyna-Q (DDQ) by combining Dyna-Q with deep learning approaches to representing the state-action space by neural networks (NN). 

By employing the world model for planning, the DDQ method can be viewed as a model-based RL approach, which has drawn growing interest in the research community. However, most model-based RL methods~\cite{tamar2016value,silver2016predictron,gu2016continuous,racaniere2017imagination} are developed for simulation-based, synthetic problems (e.g., games), but not for human-in-the-loop, real-world problems. To these ends, our main contributions in this work are two-fold:
\begin{compactitem}
\item We present Deep Dyna-Q, which to the best of our knowledge is the first deep RL framework that incorporates planning for task-completion dialogue policy learning.
\item We demonstrate that a task-completion dialogue agent can efficiently adapt its policy on the fly, by interacting with real users via RL. This results in a significant improvement in success rate on a nontrivial task.
\end{compactitem}

\section{Dialogue Policy Learning via Deep Dyna-Q (DDQ)}
\label{sec:method}
\begin{figure}[t]
\centering
\includegraphics[width=1.0\linewidth]{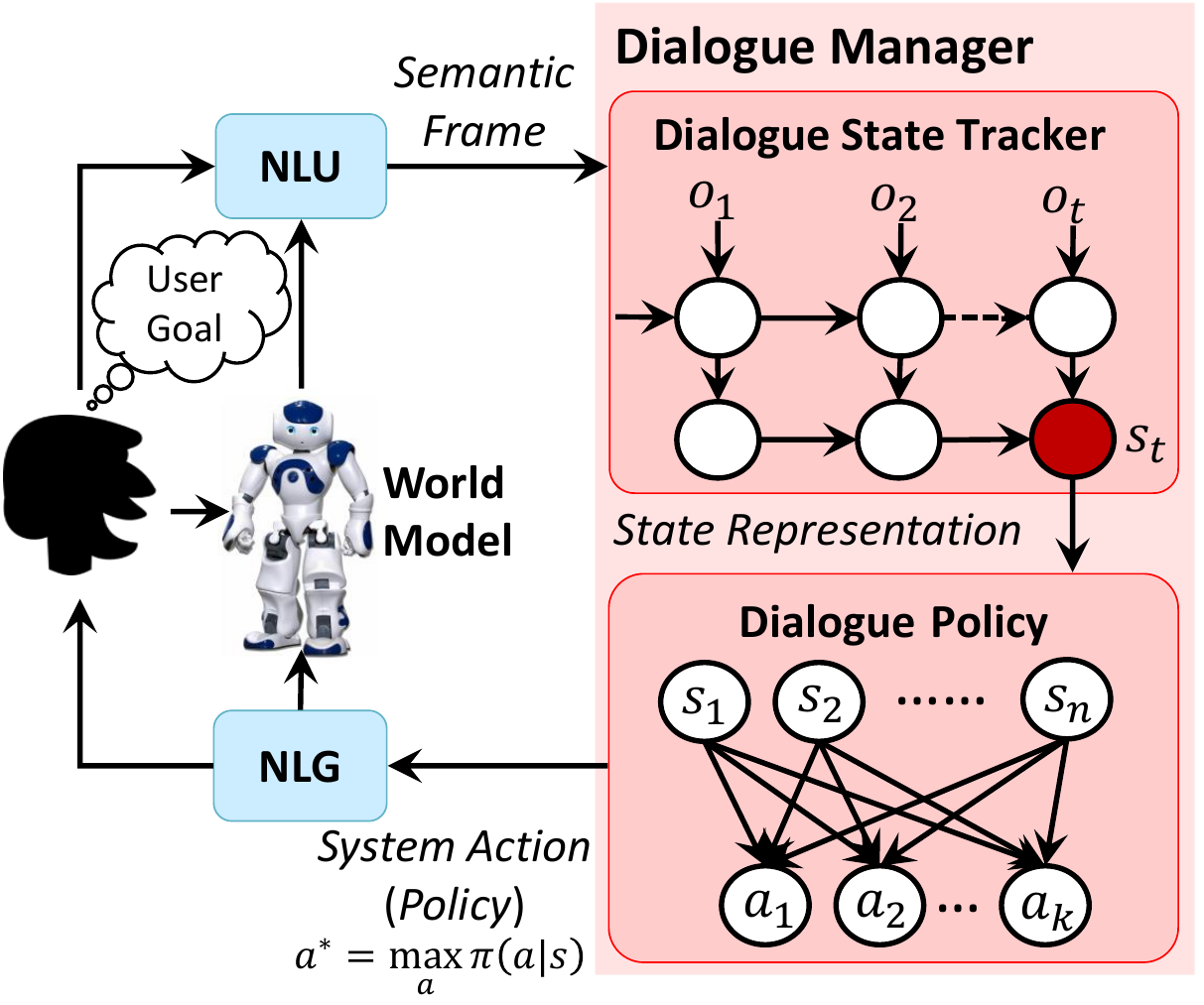}
\vspace{-5mm}
\caption{Illustration of the task-completion DDQ dialogue agent.}
\label{fig:dialogue_arch}
\end{figure}

Our DDQ dialogue agent is illustrated in Figure~\ref{fig:dialogue_arch}, consisting of five modules: (1) an LSTM-based natural language understanding (NLU) module~\cite{hakkani2016multi} for identifying user intents and extracting associated slots; (2) a state tracker~\cite{mrkvsic2016neural} for tracking the dialogue states; (3) a dialogue policy which selects the next action\footnote{In the dialogue scenario, actions are dialogue-acts, consisting of a single act and a (possibly empty) collection of $(slot=value)$ pairs~\cite{schatzmann2007agenda}.} based on the current state; (4) a model-based natural language generation (NLG) module for converting dialogue actions to natural language response~\cite{wen2015semantically}; and (5) a world model for generating simulated user actions and simulated rewards.

As illustrated in Figure~\ref{fig:learning_typeC}, starting with an initial dialogue policy and an initial world model (both trained with pre-collected human conversational data), the training of the DDQ agent consists of three processes: (1) \emph{direct reinforcement learning}, where the agent interacts with a real user, collects real experience and improves the dialogue policy; (2) \emph{world model learning}, where the world model is learned and refined using real experience; and (3) \emph{planning}, where the agent improves the dialogue policy using simulated experience.

Although these three processes conceptually can occur simultaneously in the DDQ agent, we implement an iterative training procedure, as shown in Algorithm~\ref{algo:rl_planner}, where we specify the order in which they occur within each iteration. In what follows, we will describe these processes in details.

\subsection{Direct Reinforcement Learning}
In this process (lines 5-18 in Algorithm~\ref{algo:rl_planner}) we use the DQN method~\cite{mnih2015human} to improve the dialogue policy based on real experience. We consider task-completion dialogue as a Markov Decision Process (MDP), where the agent interacts with a user in a sequence of actions to accomplish a user goal. In each step, the agent observes the dialogue state $s$, and chooses the action $a$ to execute, using an $\epsilon$-greedy policy that selects a random action with probability $\epsilon$ or otherwise follows the greedy policy $a=\text{argmax}_{a'} Q(s,a';\theta_{Q})$. $Q(s,a;\theta_{Q})$ which is the approximated value function, implemented as a Multi-Layer Perceptron (MLP) parameterized by $\theta_{Q}$. The agent then receives reward\footnote{In the dialogue scenario, reward is defined to measure the degree of success of a dialogue. In our experiment, for example, success corresponds to a reward of $80$, failure to a reward of $-40$, and the agent receives a reward of $-1$ at each turn so as to encourage shorter dialogues.} $r$, observes next user response $a^u$, and updates the state to $s'$. Finally, we store the experience $(s, a, r, a^u, s')$ in the replay buffer $D^u$. The cycle continues until the dialogue terminates. 

We improve the value function $Q(s,a;\theta_{Q})$ by adjusting $\theta_{Q}$ to minimize the mean-squared loss function, defined as follows:
\begin{eqnarray}
\mathcal{L}(\theta_{Q})&=&\mathbb{E}_{(s,a,r,s')\sim \mathcal{D}^u}[(y_i - Q(s, a;\theta_{Q}))^2] \nonumber \\
 y_i&=&r +\gamma \max_{a'}Q'(s', a';\theta_{Q'})\,
\end{eqnarray}
where $\gamma \in [0,1]$ is a discount factor, and $Q'(.)$ is the target value function that is only periodically updated (line 42 in Algorithm \ref{algo:rl_planner}). By differentiating the loss function with respect to $\theta_{Q}$, we arrive at the following gradient:
\begin{eqnarray}
\label{eq:gradient}
\begin{split}
\nabla_{\theta_{Q}}\mathcal{L}(\theta_{Q}) &= \mathbb{E}_{(s,a,r,s')\sim \mathcal{D}^u} [(r + \\
\gamma \max_{a'} Q'&(s',a';\theta_{Q'}) - Q(s,a;\theta_{Q})) \\
  & \nabla_{\theta_{Q}}Q(s,a;\theta_{Q})]
\end{split}
\end{eqnarray}
As shown in lines 16-17 in Algorithm \ref{algo:rl_planner}, in each iteration, we improve $Q(.)$ using minibatch Deep Q-learning.

\begin{algorithm}[!ht]
\small
\caption{Deep Dyna-Q for Dialogue Policy Learning}
\begin{algorithmic}[1]
\REQUIRE $N$, $\epsilon$, $K$, $L$, $C$, $Z$
\ENSURE $Q(s,a;\theta_{Q})$, $M(s,a;\theta_{M})$

\STATE initialize $Q(s,a;\theta_{Q})$ and $M(s,a;\theta_{M})$ via pre-training on human conversational data
\STATE initialize $Q'(s,a;\theta_{Q'})$ with $\theta_{Q'} = \theta_{Q}$
\STATE initialize real experience replay buffer $D^u$ using Reply Buffer Spiking (RBS), and simulated experience replay buffer $D^s$ as empty

\FOR {$n$=$1$:$N$}

\STATE \COMMENT{\textit{Direct Reinforcement Learning} starts}
\STATE \textit{user} starts a dialogue with user action $a^{u}$
\STATE generate an initial dialogue state $s$
\WHILE{$s$ is not a terminal state}
\STATE with probability $\epsilon$ select a random action $a$
\STATE otherwise select $a = \text{argmax}_{a'} Q(s,a';\theta_{Q})$
\STATE execute $a$, and observe \textit{user} response $a^u$ and reward $r$
\STATE update dialogue state to $s'$
\STATE store $(s, a, r, a^{u}, s')$ to $D^u$
\STATE $s = s'$
\ENDWHILE
\STATE sample random minibatches of $(s, a, r, s')$ from $D^u$
\STATE update $\theta_{Q}$ via $Z$-step minibatch Q-learning according to Equation~(\ref{eq:gradient})
\STATE \COMMENT{\textit{Direct Reinforcement Learning} ends}

\STATE \COMMENT{\textit{World Model Learning} starts}
\STATE sample random minibatches of training samples $(s, a, r, a^{u}, s')$ from $D^u$
\STATE update $\theta_{M}$ via $Z$-step minibatch SGD of multi-task learning
\STATE \COMMENT{\textit{World Model Learning} ends}

\STATE \COMMENT{\textit{Planning} starts}
\FOR {$k$=$1$:$K$}
\STATE $t$ = FALSE, $l=0$
\STATE sample a user goal $G$ 
\STATE sample user action $a^{u}$ from $G$
\STATE generate an initial dialogue state $s$
\WHILE{$t$ is FALSE $\land$ $l\leq L$} 
\STATE with probability $\epsilon$ select a random action $a$
\STATE otherwise select $a = \text{argmax}_{a'} Q(s,a';\theta_{Q})$
\STATE execute $a$
\STATE \textit{world model} responds with $a^u$, $r$ and $t$
\STATE update dialogue state to $s'$
\STATE store $(s, a, r, s')$ to $D^s$
\STATE $l=l+1$, $s = s'$
\ENDWHILE
\STATE sample random minibatches of $(s, a, r, s')$ from $D^s$
\STATE update $\theta_{Q}$ via $Z$-step minibatch Q-learning according to Equation~(\ref{eq:gradient})
\ENDFOR
\STATE \COMMENT{\textit{Planning} ends}

\STATE every $C$ steps reset $\theta_{Q'} = \theta_{Q}$
\ENDFOR
\end{algorithmic}
\label{algo:rl_planner}
\vspace{-1mm}
\end{algorithm}

\subsection{Planning}
In the planning process (lines 23-41 in Algorithm~\ref{algo:rl_planner}), the world model is employed to generate simulated experience that can be used to improve dialogue policy. 
$K$ in line 24 is the number of planning steps that the agent performs per step of direct reinforcement learning. If the world model is able to accurately simulate the environment, a big $K$ can be used to speed up the policy learning. In DDQ, we use two replay buffers, $D^u$ for storing real experience and $D^s$ for simulated experience. Learning and planning are accomplished by the same DQN algorithm, operating on real experience in $D^u$ for learning and on simulated experience in $D^s$ for planning. Thus, here we only describe the way the simulated experience is generated.

Similar to~\citet{schatzmann2007agenda}, at the beginning of each dialogue, we uniformly draw a user goal $G=(C,R)$, where $C$ is a set of constraints and $R$ is a set of requests (line 26 in Algorithm~\ref{algo:rl_planner}). For movie-ticket booking dialogues, constraints are typically the name and the date of the movie, the number of tickets to buy, etc. Requests can contain these slots as well as the location of the theater, its start time, etc. Table~\ref{tab:sample_dialogues_comparison} presents some sampled user goals and dialogues generated by simulated and real users, respectively. The first user action $a^u$ (line 27) can be either a request or an inform dialogue-act. A request, such as \texttt{request(theater; moviename=batman)}, consists of a request slot and multiple ($\geqslant 1$) constraint slots, uniformly sampled from $R$ and $C$, respectively. An inform contains constraint slots only. The user action can also be converted to natural language via NLG, e.g., \texttt{"which theater will show batman?"}

\begin{figure}[htb]
\centering
\vspace{-2mm}
\includegraphics[clip=true, trim=1.65cm 18.7cm 12cm 3.3cm, width=1\linewidth, scale=1.0]{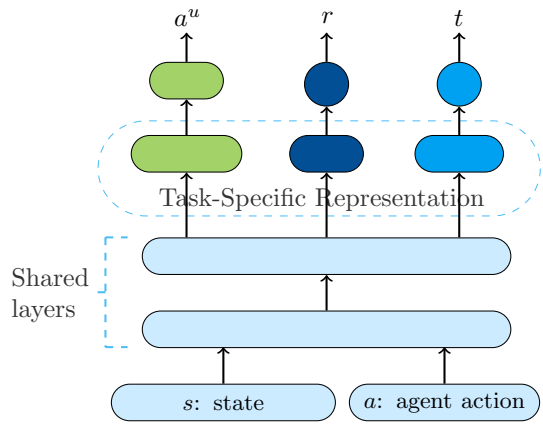}
\vspace{-6mm}
\caption{The world model architecture.}
\label{fig:planner_arch}
\vspace{-2mm}
\end{figure}

In each dialogue turn, the world model takes as input the current dialogue state $s$ and the last agent action $a$ (represented as an one-hot vector), and generates user response $a^u$, reward $r$, and a binary variable $t$, which indicates whether the dialogue terminates (line 33). The generation is accomplished using the world model $M(s, a; \theta_{M})$, a MLP shown in Figure~\ref{fig:planner_arch}, as follows:
\begin{eqnarray*}
h &=& \text{tanh} (W_{h} (s, a) + b_h)\\
r &=& W_r h + b_r \\
a^u &=& \text{softmax} (W_a h + b_a) \\
t &=& \text{sigmoid} (W_t h + b_t)
\end{eqnarray*}
where $(s,a)$ is the concatenation of $s$ and $a$, and $W$ and $b$ are parameter matrices and vectors, respectively.

\subsection{World Model Learning}
In this process (lines 19-22 in Algorithm~\ref{algo:rl_planner}), $M(s,a;\theta_{M})$ is refined via minibatch SGD using real experience in the replay buffer $D^u$. As shown in Figure~\ref{fig:planner_arch}, $M(s,a;\theta_{M})$ is a multi-task neural network~\cite{liu2015representation} that combines two classification tasks of simulating $a^u$ and $t$, respectively, and one regression task of simulating $r$. The lower layers are shared across all tasks, while the top layers are task-specific.

\section{Experiments and Results}
We evaluate the DDQ method on a movie-ticket booking task in both simulation and human-in-the-loop settings. 

\subsection{Dataset}
\label{sec:dataset}
Raw conversational data in the movie-ticket booking scenario was collected via Amazon Mechanical Turk. The dataset has been manually labeled based on a schema defined by domain experts, as shown in Table~\ref{tab:data_schema}, which consists of 11 dialogue acts and 16 slots. 
In total, the dataset contains 280 annotated dialogues, the average length of which is approximately 11 turns. 

\subsection{Dialogue Agents for Comparison}
\label{sec:dialogueagents}
To benchmark the performance of DDQ, we have developed different versions of task-completion dialogue agents, using variations of Algorithm~\ref{algo:rl_planner}.

\begin{compactitem}
\item A \textbf{DQN} agent is learned by standard DQN, implemented with direct reinforcement learning only (lines 5-18 in Algorithm~\ref{algo:rl_planner}) in each epoch.
\item The \textbf{DDQ($K$)} agents are learned by DDQ of Algorithm~\ref{algo:rl_planner}, with an initial world model pre-trained on human conversational data, as described in Section~\ref{sec:dataset}. $K$ is the number of planning steps. We trained different versions of DDQ($K$) with different $K$'s.
\item The \textbf{DDQ($K$, rand-init $\theta_M$)} agents are learned by the DDQ method with a randomly initialized world model. 
\item The \textbf{DDQ($K$, fixed $\theta_M$)} agents are learned by DDQ with an initial world model pre-trained on human conversational data. But the world model is not updated afterwards. That is, the \textit{world model learning} part in Algorithm~\ref{algo:rl_planner} (lines 19-22) is removed. The DDQ($K$, fixed $\theta_M$) agents are evaluated in the simulation setting only.
\item The \textbf{DQN($K$)} agents are learned by DQN, but with $K$ times more real experiences than the DQN agent. DQN($K$) is evaluated in the simulation setting only. Its performance can be viewed as the upper bound of its DDQ($K$) counterpart, assuming that the world model in DDQ($K$) perfectly matches real users.
\end{compactitem}

\begin{table*}[t]
\small
\begin{center}
\begin{tabular}{lccccccccc}
\\ \hline
\multirow{2}{*}{Agent}& \multicolumn{3}{c}{Epoch = 100} & \multicolumn{3}{c}{Epoch = 200} & \multicolumn{3}{c}{Epoch = 300} \\ 
\cline{2-10}
 & Success & Reward & Turns & Success & Reward & Turns & Success & Reward & Turns \\ \hline
DQN & .4260 & -3.84 & 31.93  & .5308 & 10.78 & 22.72 & .6480 & 27.66 & 22.21 \\ 
DDQ(5) & .6056 & 20.35 & 26.65 & .7128 & 36.76 & 19.55 & .7372 & 39.97 & 18.99 \\ 
DDQ(5, rand-init $\theta_M$) & .5904& 18.75 & 26.21 & .6888 & 33.47 & 20.36 & .7032 & 36.06 & 18.64\\ 
DDQ(5, fixed $\theta_M$) & .5540 & 14.54 & 25.89 & .6660 & 29.72 & 22.39 & .6860 & 33.58 &	19.49 \\ 
DQN(5) & \textit{.6560}& \textit{29.38} & \textit{21.76} & \textit{.7344} & \textit{41.09} & \textit{16.07} & \textit{.7576} & \textit{43.97} & \textit15.88\\ 
DDQ(10) & \textbf{\textcolor{blue}{.6624}} & 28.18 & 24.62 & \textbf{\textcolor{blue}{.7664}} & 42.46 & 21.01 & \textbf{\textcolor{blue}{.7840}} & 45.11 & 19.94 \\
DDQ(10, rand-init $\theta_M$) & .6132 & 21.50 & 26.16 & .6864 & 32.43 & 21.86 & .7628 & 42.37 & 20.32 \\ 
DDQ(10, fixed $\theta_M$) & .5884 & 18.41 & 26.41 & .6196 & 24.17 & 22.36 & .6412 & 26.70 &	22.49 \\ 
DQN(10) & \textit{.7944} & \textit{48.61} & \textit{15.43} & \textit{.8296} & \textit{54.00} & \textit{13.09} & \textit{.8356} & \textit{54.89} & \textit{12.77} \\
\hline
\end{tabular}
\end{center}
\vspace{-2mm}
\caption{Results of different agents at training epoch = \{100, 200, 300\}. Each number is averaged over 5 runs, each run tested on 2000 dialogues. 
Excluding DQN(5) and DQN(10) which serve as the upper bounds, any two groups of success rate (except three groups: at epoch 100, DDQ(5, rand-init $\theta_M$) and DDQ(10, fixed $\theta_M$), at epoch 200, DDQ(5, rand-init $\theta_M$) and DDQ(10, rand-init $\theta_M$), at epoch 300, DQN and DDQ(10, fixed $\theta_M$))
evaluated at the same epoch is statistically significant in mean with $p<0.01$.
(Success: success rate)}
\label{tab:test_results}
\vspace{-2mm}
\end{table*}

\paragraph{Implementation Details} All the models in these agents ($Q(s,a;\theta_{Q})$, $M(s,a;\theta_{M})$) are MLPs with \texttt{tanh} activations. Each policy network $Q(.)$ has one hidden layer with 80 hidden nodes. As shown in Figure~\ref{fig:planner_arch}, the world model $M(.)$ contains two shared hidden layers and three task-specific hidden layers, with 80 nodes in each. All the agents are trained by Algorithm~\ref{algo:rl_planner} with the same set of hyper-parameters. $\epsilon$-greedy is always applied for exploration. We set the discount factor $\gamma$ = 0.95. The buffer sizes of both $D^u$ and $D^s$ are set to 5000. The target value function is updated at the end of each epoch. In each epoch, $Q(.)$ and $M(.)$ are refined using one-step ($Z=1$) $16$-tuple-minibatch update.~\footnote{We found in our experiments that setting $Z>1$ improves the performance of all agents, but does not change the conclusion of this study: DDQ consistently outperforms DQN by a statistically significant margin. Conceptually, the optimal value of $Z$ used in planning is different from that in direct reinforcement learning, and should vary according to the quality of the world model. The better the world model is, the more aggressive update (thus bigger $Z$) is being used in planning. We leave it to future work to investigate how to optimize $Z$ for planning in DDQ.} In planning, the maximum length of a simulated dialogue is 40 ($L=40$). In addition, to make the dialogue training efficient, we also applied a variant of imitation learning, called Reply Buffer Spiking (RBS)~\cite{lipton2016efficient}. We built a naive but occasionally successful rule-based agent based on human conversational dataset (line 1 in Algorithm~\ref{algo:rl_planner}), and pre-filled the real experience replay buffer $D^u$ with 100 dialogues of experience (line 2) before training for all the variants of agents. 

\begin{figure}[t]
\centering  
\vspace{-2mm}
\includegraphics[width=1\linewidth]{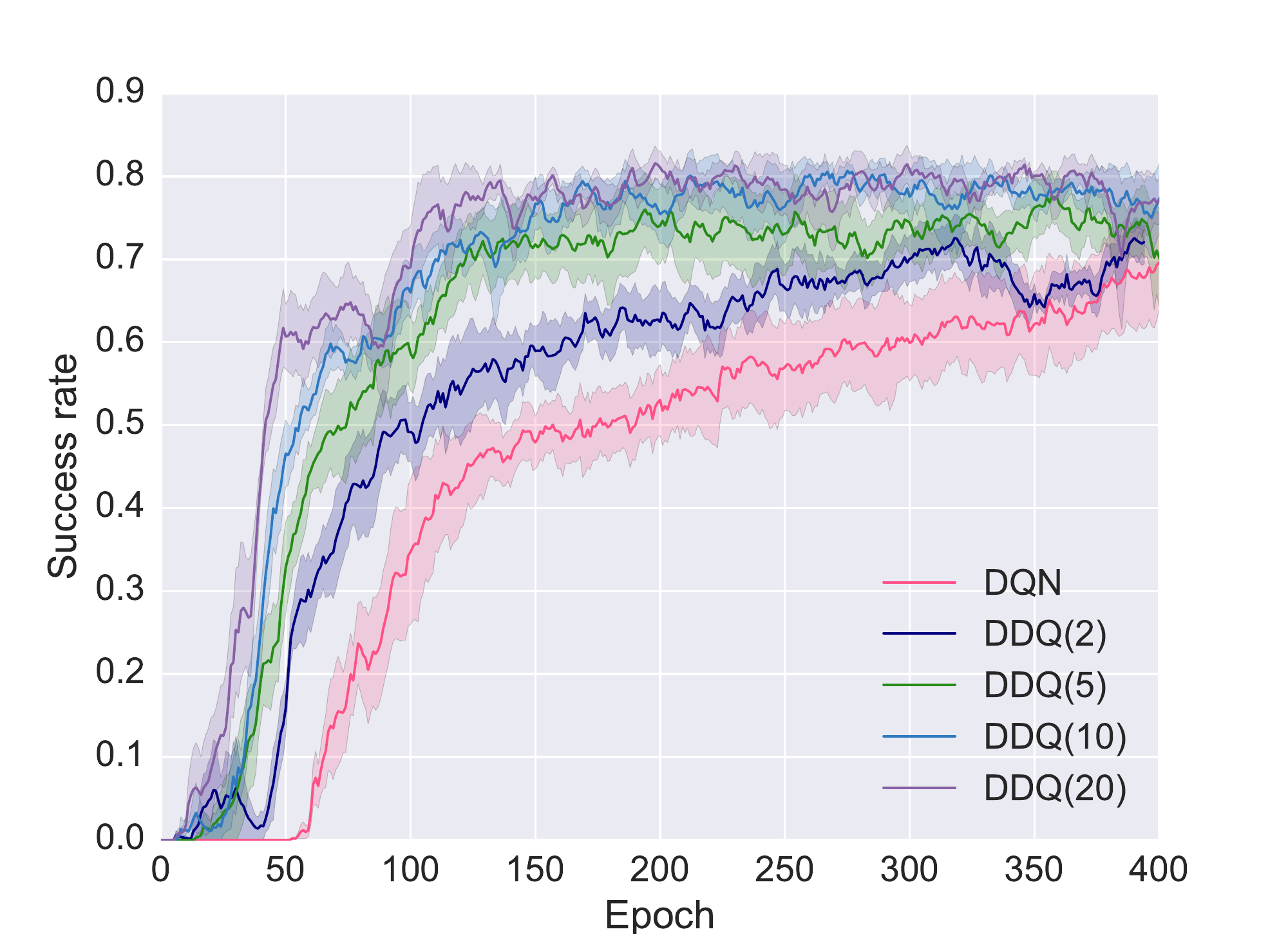}
\vspace{-6mm}
\caption{Learning curves of the DDQ($K$) agents with $K=2,5,10,20$. The DQN agent is identical to a DDQ($K$) agent with $K=0$.}
\label{fig:exp_s3} 
\vspace{-2mm}
\end{figure}


\subsection{Simulated User Evaluation}
\label{sec:simulated_user_evaluation}
In this setting the dialogue agents are optimized by interacting with user simulators, instead of real users. Thus, the world model is learned to mimic user simulators. Although the simulator-trained agents are sub-optimal when applied to real users due to the discrepancy between simulators and real users, the simulation setting allows us to perform a detailed analysis of DDQ without much cost and to reproduce the experimental results easily. 

\paragraph{User Simulator} We adapted a publicly available user simulator~\cite{li2016user} to the task-completion dialogue setting. 
During training, the simulator provides the agent with a simulated user response in each dialogue turn and a reward signal at the end of the dialogue. A dialogue is considered successful only when a movie ticket is booked successfully and when the information provided by the agent satisfies all the user's constraints. At the end of each dialogue, the agent receives a positive reward of $2*L$ for success, or a negative reward of $-L$ for failure, where $L$ is the maximum number of turns in each dialogue, and is set to $40$ in our experiments. Furthermore, in each turn, the agent receives a reward of $-1$, so that shorter dialogues are encouraged. Readers can refer to Appendix~\ref{app:user_sim} for details on the user simulator.


\begin{figure}[t]
\centering  
\vspace{-2mm}
\includegraphics[width=1\columnwidth]{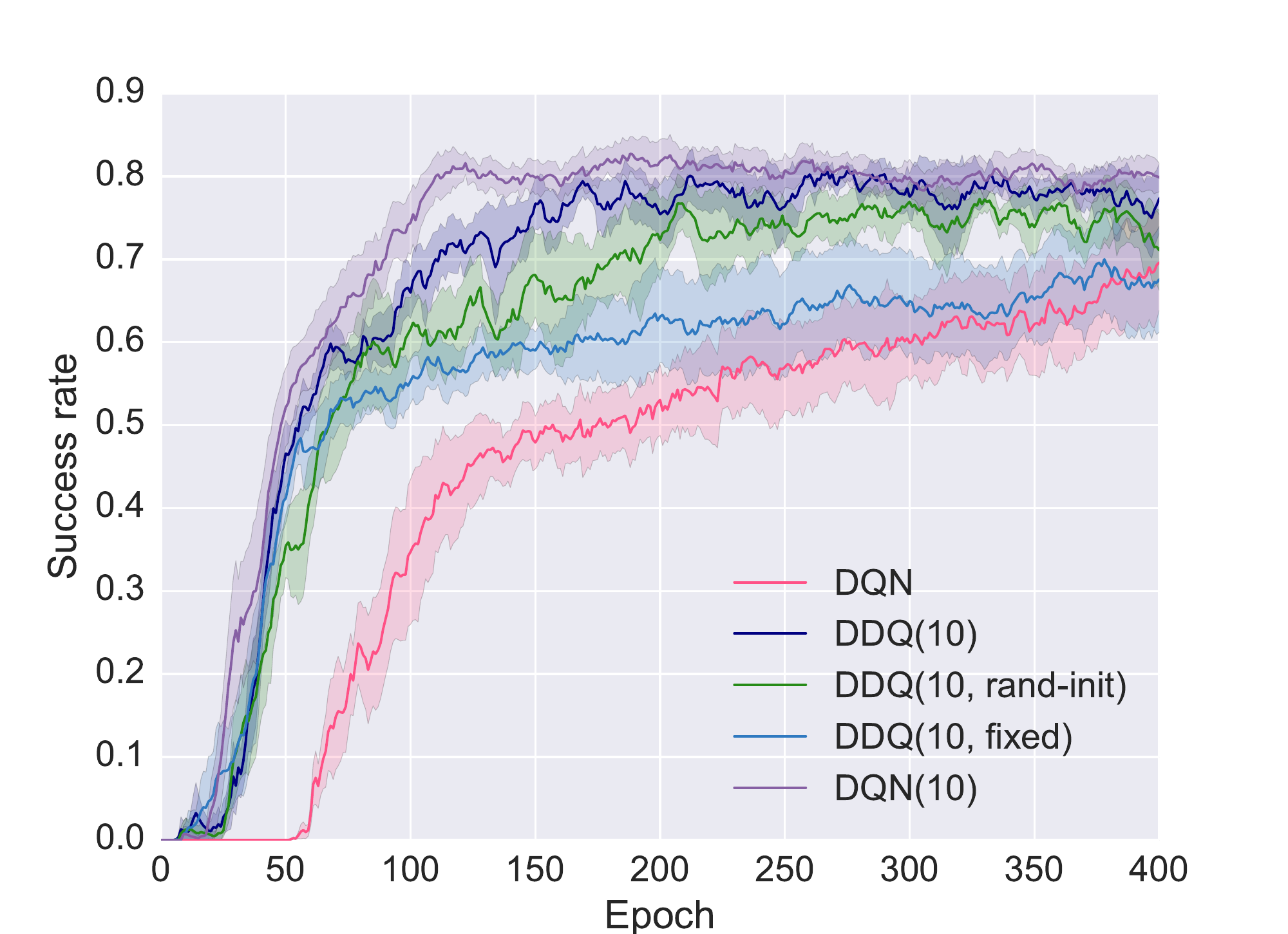}
\vspace{-6mm}
\caption{Learning curves of DQN, DDQ(10), DDQ(10, rand-init $\theta_M$), DDQ(10, fixed $\theta_M$), and DQN(10).} 
\label{fig:exp_s4} 
\vspace{-2mm}
\end{figure}

\begin{table*}[t]
\small
\begin{center}
\begin{tabular}{lccccccccc}
\\ \hline
\multirow{2}{*}{Agent}& \multicolumn{3}{c}{Epoch = 100} & \multicolumn{3}{c}{Epoch = 150} & \multicolumn{3}{c}{Epoch = 200} \\ 
\cline{2-10}
 & Success & Reward & Turns & Success & Reward & Turns & Success & Reward & Turns \\ \hline
DQN & .0000 & -58.69 & 39.38  & .4080 & -5.730 & 30.38 & .4545 & 0.350 & 30.38 \\ 
DDQ(5) & .4620 & 00.78 & 31.33 & .5637 & 15.05 & 26.17 & .6000 & 19.84 & 26.32\\ 
DDQ(5, rand-init $\theta_M$) & .3600 & -11.67 & 31.74 & .5500 & 13.71 & 26.58 & .5752 & 16.84 & 26.37 \\
DDQ(10) & \textbf{\textcolor{blue}{.5555}} & 14.69 & 25.92 & \textbf{\textcolor{blue}{.6416}} & 25.85 & 24.28 & \textbf{\textcolor{blue}{.7332}} & 38.88 & 20.21 \\ 
DDQ(10, rand-init $\theta_M$) & .5010 & 6.27 & 29.70 & .6055 & 22.11 & 23.11 & .7023 & 36.90 & 21.20 \\ 
\hline
\end{tabular}
\end{center}
\vspace{-2mm}
\caption{The performance of different agents at training epoch = \{100, 150, 200\} in the human-in-the-loop experiments. The difference between the results of all agent pairs evaluated at the same epoch is statistically significant ($p<0.01$).
(Success: success rate)}
\label{tab:human_test_results}
\vspace{-2mm}
\end{table*}

\paragraph{Results} 
The main simulation results
are reported in Table~\ref{tab:test_results} and Figures~\ref{fig:exp_s3} and~\ref{fig:exp_s4}. For each agent, we report its results in terms of success rate, average reward, and average number of turns (averaged over 5 repetitions of the experiments). 
Results show that the DDQ agents consistently outperform DQN with a statistically significant margin. Figure~\ref{fig:exp_s3} shows the learning curves of different DDQ agents trained using different planning steps. Since the training of all RL agents started with RBS using the same rule-based agent, their performance in the first few epochs is very close. After that, performance improved for all values of $K$, but much more rapidly for larger values. Recall that the DDQ($K$) agent with $K$=0 is identical to the DQN agent, which does no planning but relies on direct reinforcement learning only. Without planning, the DQN agent took about 180 epochs (real dialogues) to reach the success rate of 50\%, and DDQ(10) took only 50 epochs. 

Intuitively, the optimal value of $K$ needs to be determined by seeking the best trade-off between the quality of the world model and the amount of simulated experience that is useful for improving the dialogue agent. This is a non-trivial optimization problem because both the dialogue agent and the world model are updated constantly during training and the optimal $K$ needs to be adjusted accordingly. For example, we find in our experiments that at the early stages of training, it is fine to perform planning aggressively by using large amounts of simulated experience even though they are of low quality, but in the late stages of training where the dialogue agent has been significantly improved, low-quality simulated experience is likely to hurt the performance. Thus, in our implementation of Algorithm~\ref{algo:rl_planner}, we use a heuristic\footnote{The heuristic is not presented in Algorithm~\ref{algo:rl_planner}. Readers can refer to the released source code for details.} to reduce the value of $K$ in the late stages of training (e.g., after 150 epochs in Figure~\ref{fig:exp_s3}) to mitigate the negative impact of low-qualify simulated experience.
We leave it to future work how to optimize the planning step size during DDQ training in a principled way.

Figure~\ref{fig:exp_s4} shows that the quality of the world model has a significant impact on the agent's performance. The learning curve of DQN(10) indicates the best performance we can expect with a \emph{perfect} world model. With a pre-trained world model, the performance of the DDQ agent improves more rapidly, although eventually, the DDQ and DDQ(rand-init $\theta_M$) agents reach the same success rate after many epochs. The world model learning process is crucial to both the efficiency of dialogue policy learning and the final performance of the agent. For example, in the early stages (before 60 epochs), the performances of DDQ and DDQ(fixed $\theta_M$) remain very close to each other, but DDQ reaches a success rate almost 10\% better than DDQ(fixed $\theta_M$) after 400 epochs.


\begin{figure}[t]
\vspace{-1mm}
\centering  
\includegraphics[width=1\linewidth]{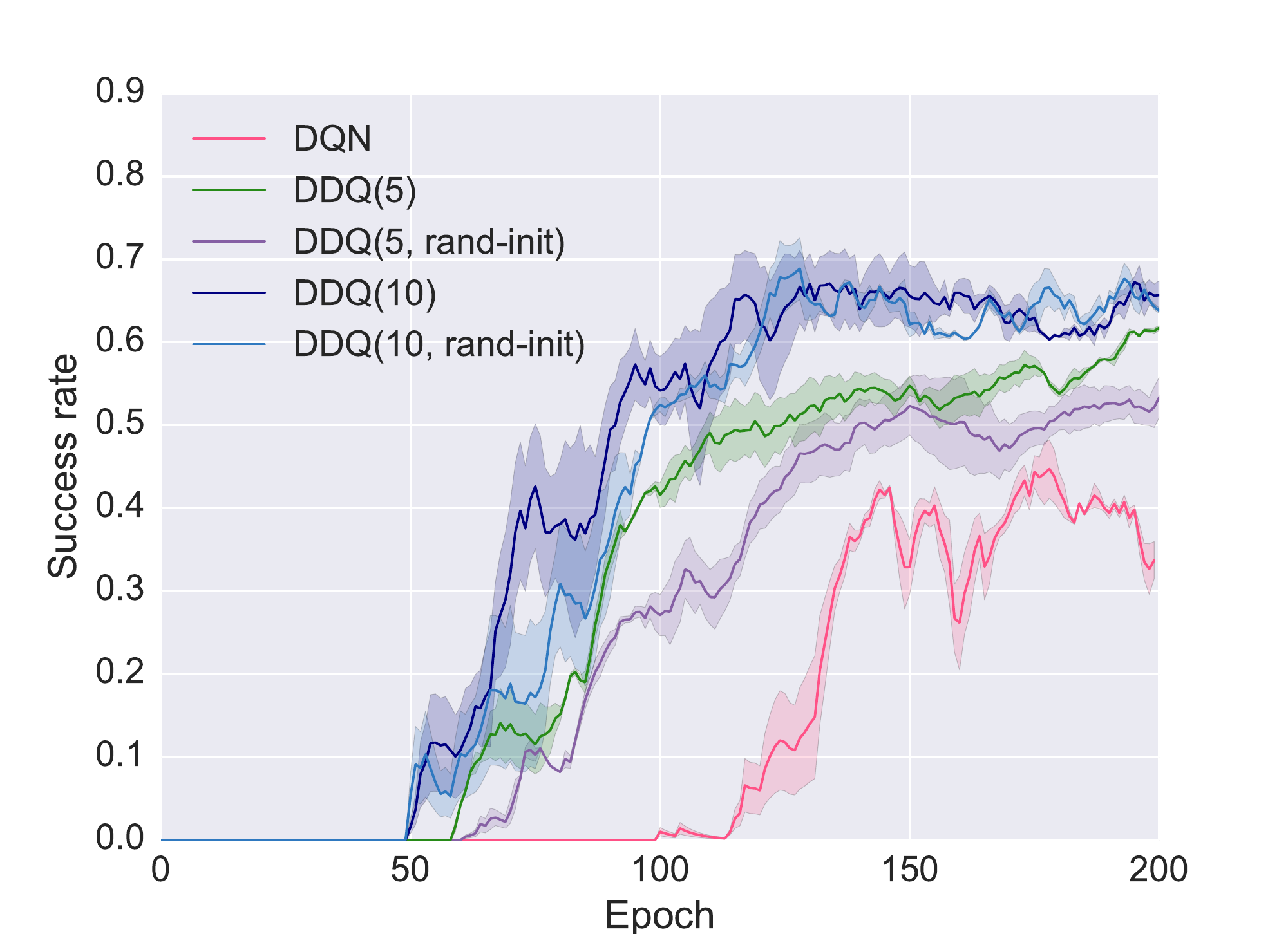}
\vspace{-6mm}
\caption{Human-in-the-loop dialogue policy learning curves in four different agents.} 
\vspace{-2mm}
\label{fig:exp_h1} 
\end{figure}

\begin{table*}[!ht]
\small
\begin{tabular}{l|l}
\hline
\multicolumn{1}{c|}{\textbf{Simulation Sample}}  & \multicolumn{1}{c}{\textbf{Real User Sample}}  \\
\hline
\begin{tabular}[c]{@{}l@{}}
\textit{movie-ticket} booking user goal:\\
\{\\
\-\hspace{1mm} ``request\_slots": \{	 \-\hspace{10mm}     ``constraint\_slots": \{\\
\-\hspace{3mm}    ``ticket": ``?"	 	 \-\hspace{17mm}    \textbf{``numberofpeople":``2"} \\
\-\hspace{3mm}    ``theater": ``?"		 \-\hspace{15mm}    \textbf{``moviename": ``deadpool"}	\\
\-\hspace{3mm}    ``starttime": ``?"	 \-\hspace{13mm}    \textbf{``city": ``seattle"} \\
\-\hspace{3mm}    ``date": ``?"	         \-\hspace{18mm}    \}\\
\-\hspace{1mm}  \} \\
\} \\
\end{tabular}  & 
\begin{tabular}[c]{@{}l@{}}
\textit{movie-ticket} booking user goal:\\
\{\\
\-\hspace{1mm} ``request\_slots": \{	 \-\hspace{10mm}     ``constraint\_slots": \{\\
\-\hspace{3mm}    ``ticket": ``?"	 	 \-\hspace{17mm}    \textbf{``date":``this weekend"} \\
\-\hspace{3mm}    ``theater": ``?"		 \-\hspace{15mm}    \textbf{``numberofpeople": ``1"}	\\
\-\hspace{3mm}    ``starttime": ``?"	 \-\hspace{13mm}      \textbf{``moviename": ``batman"} \\
\-\hspace{1mm}  \}	 \-\hspace{32mm}       \}\\
\} \\
\\
\end{tabular}                \\
\hline
\multicolumn{2}{c}{\textbf{DQN agent}} \\

\begin{tabular}[c]{@{}p{4mm}@{}|l@{}} 1 & \textit{usr}: Hi, what date can I book 2 tickets for deadpool? \end{tabular} & \textit{usr}: Hi, which theater will show batman? I want 1 ticket.  \\
\begin{tabular}[c]{@{}p{4mm}@{}|l@{}} 2 & \textit{agt}: Which city would you like? \end{tabular} & \textit{agt}: Which city would you like?  \\
\begin{tabular}[c]{@{}p{4mm}@{}|l@{}} 3 & \textit{usr}: I want to watch at seattle. \end{tabular} & \textit{usr}: Boston.  \\
\begin{tabular}[c]{@{}p{4mm}@{}|l@{}} 4 & \textit{agt}: What time would you like to see it? \end{tabular} & \textit{agt}: What time would you like to see it? \\
\begin{tabular}[c]{@{}p{4mm}@{}|l@{}} 5 & \textit{usr}: What start time is available? \end{tabular} & \textit{usr}: What start time is available? \\
\begin{tabular}[c]{@{}p{4mm}@{}|l@{}} 6 & \textit{agt}: \textcolor{blue}{\underline{Which theater would you like?}} \end{tabular} & \textit{agt}: What date would you like to see it?  \\
\begin{tabular}[c]{@{}p{4mm}@{}|l@{}} 7 & \textit{usr}: Which theater is available? \end{tabular} & \textit{usr}: This weekend. \\
\begin{tabular}[c]{@{}p{4mm}@{}|l@{}} 8 &\textit{agt}: What time would you like to see it? \end{tabular} &  \textit{agt}:\textcolor{blue}{\underline{Which theater would you like?}} \\
\begin{tabular}[c]{@{}p{4mm}@{}|l@{}} 9 &\textit{usr}: What start time is available? \end{tabular} & \textit{usr}: Which theater is available?  \\   
\begin{tabular}[c]{@{}p{4mm}@{}|l@{}} 10 &\textit{agt}: \textcolor{blue}{\underline{Which theater would you like?}}  \end{tabular} & \textit{agt}: \textcolor{blue}{\underline{Which theater would you like?}}  \\
\begin{tabular}[c]{@{}p{4mm}@{}|l@{}} 11 &\textit{usr}: Which theater is available?	\end{tabular} &  \textit{usr}: Which theater is available?      \\
\begin{tabular}[c]{@{}p{4mm}@{}|l@{}} 12 &\textit{agt}: How many tickets do you need?  \end{tabular} &  \textit{agt}: What time would you like to see it?   \\
\begin{tabular}[c]{@{}p{4mm}@{}|l@{}} 13 &\textit{usr}: I want 2 tickets please! \end{tabular} & \textit{usr}: What start time is available? \\ 
\begin{tabular}[c]{@{}p{4mm}@{}|l@{}} 14 &\textit{agt}: \textcolor{blue}{\underline{Which theater would you like?}} \end{tabular} & \textit{agt}: \textcolor{blue}{\underline{Which theater would you like?}}  \\
\begin{tabular}[c]{@{}p{4mm}@{}|l@{}} 15 &\textit{usr}: Which theater is available? \end{tabular} & \textit{usr}: Thanks, I do not need the ticket. \\
\begin{tabular}[c]{@{}p{4mm}@{}|l@{}} 16 &\textit{agt}: What time would you like to see it? \end{tabular} & \textit{agt}: Thank you. \\
\multicolumn{1}{c|}{\textbf{\vdots}} &  \\
\begin{tabular}[c]{@{}p{4mm}@{}|l@{}} 39 &\textit{agt}: \textcolor{blue}{\underline{Which theater would you like?}} \end{tabular} &   \\
\begin{tabular}[c]{@{}p{4mm}@{}|l@{}} 40 &\textit{usr}: Which theater is available? \end{tabular} &   \\
\hline
\textbf{Failure:} \begin{tabular}[c]{@{}l@{}}Agent does not book the right ticket within the \\ max turns.\end{tabular} & \textbf{Failure:} \begin{tabular}[c]{@{}l@{}} User terminates the dialogue before reaching the \\ max turns because the agent repeatly asks the same \\ question. \end{tabular}
\\ \hline

\multicolumn{2}{c}{\textbf{DDQ(10) agent}} \\

\begin{tabular}[c]{@{}p{4mm}@{}|l@{}} 1 & \textit{usr}: Hi, when is deadpool playing at seattle? \end{tabular} & \textit{usr}: Hi, which theater will show batman? I want 1 ticket. \\
\begin{tabular}[c]{@{}p{4mm}@{}|l@{}} 2 & \textit{agt}: Tonight.  \end{tabular}  & \textit{agt}: Century Eastport 16.  \\
\begin{tabular}[c]{@{}p{4mm}@{}|l@{}} 3 & \textit{usr}: What is the start time?  \end{tabular}  & \textit{usr}: What is the start time? \\
\begin{tabular}[c]{@{}p{4mm}@{}|l@{}} 4 & \textit{agt}: Which theater would you like?  \end{tabular}  & \textit{agt}: 9PM.  \\
\begin{tabular}[c]{@{}p{4mm}@{}|l@{}} 5 & \textit{usr}: I want to watch it at amc pacific place 11.  \end{tabular} & \textit{usr}: Can I buy a ticket here?  \\
\begin{tabular}[c]{@{}p{4mm}@{}|l@{}} 6 & \textit{agt}: 10pm is available.  \end{tabular}   & \textit{agt}: This weekend?  \\
\begin{tabular}[c]{@{}p{4mm}@{}|l@{}} 7 & \textit{usr}: I want 2 tickets please!  \end{tabular} & \textit{usr}: Yes, one ticket please! \\
\begin{tabular}[c]{@{}p{4mm}@{}|l@{}} 8 & \textit{agt}: \begin{tabular}[c]{@{}l@{}} Great, here are 2 deadpool tickets for you \\ at amc pacific place 11 in seattle tonight 10 pm. \end{tabular} \end{tabular}  &  \textit{agt}: \begin{tabular}[c]{@{}l@{}} Sure, here it is - batman at Century Eastport 16, \\ 9PM this weekend. \end{tabular}      \\
\begin{tabular}[c]{@{}p{4mm}@{}|l@{}} 9 & \textit{usr}: Thank you. \end{tabular}   &    \textit{usr}: Thank you.  \\   
\begin{tabular}[c]{@{}p{4mm}@{}|l@{}} 10 &\textit{agt}: Thank you. \end{tabular}   &    \textit{agt}: Thank you.  \\   
\hline
\multicolumn{1}{c|}{\textbf{Success}}  & \multicolumn{1}{c}{\textbf{Success}}
\\ \hline
\end{tabular}
\small
\centering
\caption{Two sample dialogue sessions by DQN and DDQ(10) agents trained at epoch 100: Left: simulated user experiments; Right: human-in-the-loop experiments. (\textit{agt}: agent, \textit{usr}: user) }
\label{tab:sample_dialogues_comparison}
\vspace{-1mm}
\end{table*}



\subsection{Human-in-the-Loop Evaluation}
\label{sec:human-in-the-loop-exp}
In this setting, five dialogue agents (i.e., DQN, DDQ(10), DDQ(10, rand-init $\theta_M$), DDQ(5), and DDQ(5, rand-init $\theta_M$)) are trained via RL by interacting with real human users. In each dialogue session, one of the agents was randomly picked to converse with a user. The user was presented with a user goal sampled from the corpus, and was instructed to converse with the agent to complete the task. 
The user had the choice of abandoning the task and ending the dialogue at any time, if she or he believed that the dialogue was unlikely to succeed or simply because the dialogue dragged on for too many turns. In such cases, the dialogue session is considered failed. At the end of each session, the user was asked to give explicit feedback whether the dialogue succeeded (i.e., whether the movie tickets were booked with all the user constraints satisfied). Each learning curve is trained with two runs, with each run generating 150 dialogues
(and $K*150$ additional simulated dialogues when planning is applied). In total, we collected 1500 dialogue sessions for training all five agents.

The main results are presented in Table~\ref{tab:human_test_results} and Figure~\ref{fig:exp_h1}, with each agent averaged over two independent runs. The results confirm what we observed in the simulation experiments. The conclusions are summarized as below:
\begin{compactitem}
\item The DDQ agent significantly outperforms DQN, as demonstrated by the comparison between DDQ(10) and DQN. Table~\ref{tab:sample_dialogues_comparison} presents four example dialogues produced by two dialogue agents interacting with simulated and human users, respectively. The DQN agent, after being trained with 100 dialogues, still behaved like a naive rule-based agent that requested information bit by bit in a fixed order. When the user did not answer the request explicitly (e.g., \texttt{usr: which theater is available?}), the agent failed to respond properly. On the other hand, with planning, the DDQ agent trained with 100 real dialogues is much more robust and can complete 50\% of user tasks successfully.
\item A larger $K$ leads to more aggressive planning and better results, as shown by DDQ(10) vs. DDQ(5).
\item Pre-training world model with human conversational data improves the learning efficiency and the agent's performance, as shown by DDQ(5) vs. DDQ(5, rand-init $\theta_M$), and DDQ(10) vs. DDQ(10, rand-init $\theta_M$).
\end{compactitem}

\section{Conclusion}
\label{sec:conclusion}
We propose a new strategy for a task-completion dialogue agent to learn its policy by interacting with real users. Compared to previous work, our agent learns in a much more efficient way, using only a small number of real user interactions, which amounts to an affordable cost in many nontrivial domains. Our strategy is based on the Deep Dyna-Q (DDQ) framework where planning is integrated into dialogue policy learning. The effectiveness of DDQ is validated by human-in-the-loop experiments, demonstrating 
that a dialogue agent can efficiently adapt its policy on the fly by interacting with real users via deep RL. 


One interesting topic for future research is \emph{exploration in planning}. We need to deal with the challenge of adapting the world model in a changing environment, as exemplified by the domain extension problem~\cite{lipton2016efficient}. As pointed out by~\citet{sutton1998introduction}, the general problem here is a particular manifestation of the conflict between exploration and exploitation. In a planning context, exploration means trying actions that may improve the world model, whereas exploitation means trying to behave in the optimal way given the current model. To this end, we want the agent to explore in the environment, but not so much that the performance would be greatly degraded.



\section*{Acknowledgments}
We would like to thank Chris Brockett, Yun-Nung Chen, Michel Galley and Lihong Li for their insightful comments on the paper. We would like to acknowledge the volunteers from Microsoft Research for helping us with the human-in-the-loop experiments. This work was done when Baolin Peng and Shang-Yu Su were visiting Microsoft. Baolin Peng is in part supported by Innovation and Technology Fund (6904333), and General Research Fund of Hong Kong (12183516).


\bibliography{acl2018}

\begin{thebibliography}{}
\expandafter\ifx\csname natexlab\endcsname\relax\def\natexlab#1{#1}\fi

\bibitem[{Budzianowski et~al.(2017)Budzianowski, Ultes, Su, Mrksic, Wen,
  Casanueva, Rojas-Barahona, and Gasic}]{budzianowski2017sub}
Pawel Budzianowski, Stefan Ultes, Pei-Hao Su, Nikola Mrksic, Tsung-Hsien Wen,
  Inigo Casanueva, Lina Rojas-Barahona, and Milica Gasic. 2017.
\newblock Sub-domain modelling for dialogue management with hierarchical
  reinforcement learning.
\newblock {\em arXiv preprint arXiv:1706.06210\/} .

\bibitem[{Dhingra et~al.(2017)Dhingra, Li, Li, Gao, Chen, Ahmed, and
  Deng}]{dhingra2017towards}
Bhuwan Dhingra, Lihong Li, Xiujun Li, Jianfeng Gao, Yun-Nung Chen, Faisal
  Ahmed, and Li~Deng. 2017.
\newblock Towards end-to-end reinforcement learning of dialogue agents for
  information access.
\newblock In {\em Proceedings of the 55th Annual Meeting of the Association for
  Computational Linguistics (Volume 1: Long Papers)\/}. volume~1, pages
  484--495.

\bibitem[{Ga{\v{s}}i{\'c} et~al.(2010)Ga{\v{s}}i{\'c},
  Jur{\v{c}}{\'\i}{\v{c}}ek, Keizer, Mairesse, Thomson, Yu, and
  Young}]{gavsic2010gaussian}
Milica Ga{\v{s}}i{\'c}, Filip Jur{\v{c}}{\'\i}{\v{c}}ek, Simon Keizer,
  Fran{\c{c}}ois Mairesse, Blaise Thomson, Kai Yu, and Steve Young. 2010.
\newblock Gaussian processes for fast policy optimisation of pomdp-based
  dialogue managers.
\newblock In {\em Proceedings of the 11th Annual Meeting of the Special
  Interest Group on Discourse and Dialogue\/}. Association for Computational
  Linguistics, pages 201--204.

\bibitem[{Ga{\v{s}}i{\'c} et~al.(2011)Ga{\v{s}}i{\'c},
  Jur{\v{c}}{\'\i}{\v{c}}ek, Thomson, Yu, and Young}]{gavsic2011line}
Milica Ga{\v{s}}i{\'c}, Filip Jur{\v{c}}{\'\i}{\v{c}}ek, Blaise Thomson, Kai
  Yu, and Steve Young. 2011.
\newblock On-line policy optimisation of spoken dialogue systems via live
  interaction with human subjects.
\newblock In {\em Automatic Speech Recognition and Understanding (ASRU), 2011
  IEEE Workshop on\/}. IEEE, pages 312--317.

\bibitem[{Gu et~al.(2016)Gu, Lillicrap, Sutskever, and
  Levine}]{gu2016continuous}
Shixiang Gu, Timothy Lillicrap, Ilya Sutskever, and Sergey Levine. 2016.
\newblock Continuous deep q-learning with model-based acceleration.
\newblock In {\em International Conference on Machine Learning\/}. pages
  2829--2838.

\bibitem[{Hakkani-T{\"u}r et~al.(2016)Hakkani-T{\"u}r, Tur, Celikyilmaz, Chen,
  Gao, Deng, and Wang}]{hakkani2016multi}
Dilek Hakkani-T{\"u}r, Gokhan Tur, Asli Celikyilmaz, Yun-Nung Chen, Jianfeng
  Gao, Li~Deng, and Ye-Yi Wang. 2016.
\newblock Multi-domain joint semantic frame parsing using bi-directional
  {RNN-LSTM}.
\newblock In {\em Proceedings of The 17th Annual Meeting of the International
  Speech Communication Association\/}.

\bibitem[{Kuvayev and Sutton(1996)}]{kuvayev1996model}
Leonid Kuvayev and Richard~S Sutton. 1996.
\newblock Model-based reinforcement learning with an approximate, learned
  model.
\newblock In {\em in Proceedings of the Ninth Yale Workshop on Adaptive and
  Learning Systems\/}. Citeseer.

\bibitem[{Levin et~al.(1997)Levin, Pieraccini, and Eckert}]{levin1997learning}
Esther Levin, Roberto Pieraccini, and Wieland Eckert. 1997.
\newblock Learning dialogue strategies within the markov decision process
  framework.
\newblock In {\em Automatic Speech Recognition and Understanding, 1997.
  Proceedings., 1997 IEEE Workshop on\/}. IEEE, pages 72--79.

\bibitem[{Li et~al.(2016{\natexlab{a}})Li, Miller, Chopra, Ranzato, and
  Weston}]{li2016dialogue}
Jiwei Li, Alexander~H Miller, Sumit Chopra, Marc'Aurelio Ranzato, and Jason
  Weston. 2016{\natexlab{a}}.
\newblock Dialogue learning with human-in-the-loop.
\newblock {\em arXiv preprint arXiv:1611.09823\/} .

\bibitem[{Li et~al.(2016{\natexlab{b}})Li, Lipton, Dhingra, Li, Gao, and
  Chen}]{li2016user}
Xiujun Li, Zachary~C Lipton, Bhuwan Dhingra, Lihong Li, Jianfeng Gao, and
  Yun-Nung Chen. 2016{\natexlab{b}}.
\newblock A user simulator for task-completion dialogues.
\newblock {\em arXiv preprint arXiv:1612.05688\/} .

\bibitem[{Li et~al.(2017)Li, Chen, Li, Gao, and Celikyilmaz}]{li2017end}
Xuijun Li, Yun-Nung Chen, Lihong Li, Jianfeng Gao, and Asli Celikyilmaz. 2017.
\newblock End-to-end task-completion neural dialogue systems.
\newblock In {\em Proceedings of the The 8th International Joint Conference on
  Natural Language Processing\/}. pages 733--743.

\bibitem[{Lipton et~al.(2016)Lipton, Gao, Li, Li, Ahmed, and
  Deng}]{lipton2016efficient}
Zachary~C Lipton, Jianfeng Gao, Lihong Li, Xiujun Li, Faisal Ahmed, and
  Li~Deng. 2016.
\newblock Efficient exploration for dialogue policy learning with bbq networks
  \& replay buffer spiking.
\newblock {\em arXiv preprint arXiv:1608.05081\/} .

\bibitem[{Liu and Lane(2017)}]{liu2017iterative}
Bing Liu and Ian Lane. 2017.
\newblock Iterative policy learning in end-to-end trainable task-oriented
  neural dialog models.
\newblock In {\em Proceedings of 2017 IEEE Workshop on Automatic Speech
  Recognition and Understanding\/}.

\bibitem[{Liu et~al.(2015)Liu, Gao, He, Deng, Duh, and
  Wang}]{liu2015representation}
Xiaodong Liu, Jianfeng Gao, Xiaodong He, Li~Deng, Kevin Duh, and Ye-Yi Wang.
  2015.
\newblock Representation learning using multi-task deep neural networks for
  semantic classification and information retrieval .

\bibitem[{Mnih et~al.(2015)Mnih, Kavukcuoglu, Silver, Rusu, Veness, Bellemare,
  Graves, Riedmiller, Fidjeland, Ostrovski et~al.}]{mnih2015human}
Volodymyr Mnih, Koray Kavukcuoglu, David Silver, Andrei~A Rusu, Joel Veness,
  Marc~G Bellemare, Alex Graves, Martin Riedmiller, Andreas~K Fidjeland, Georg
  Ostrovski, et~al. 2015.
\newblock Human-level control through deep reinforcement learning.
\newblock {\em Nature\/} 518(7540):529--533.

\bibitem[{Moore and Atkeson(1993)}]{moore1993prioritized}
Andrew~W Moore and Christopher~G Atkeson. 1993.
\newblock Prioritized sweeping: Reinforcement learning with less data and less
  time.
\newblock {\em Machine learning\/} 13(1):103--130.

\bibitem[{Mrk{\v{s}}i{\'c} et~al.(2016)Mrk{\v{s}}i{\'c}, S{\'e}aghdha, Wen,
  Thomson, and Young}]{mrkvsic2016neural}
Nikola Mrk{\v{s}}i{\'c}, Diarmuid~O S{\'e}aghdha, Tsung-Hsien Wen, Blaise
  Thomson, and Steve Young. 2016.
\newblock Neural belief tracker: Data-driven dialogue state tracking.
\newblock {\em arXiv preprint arXiv:1606.03777\/} .

\bibitem[{Papangelis(2012)}]{papangelis2012comparative}
Alexandros Papangelis. 2012.
\newblock A comparative study of reinforcement learning techniques on dialogue
  management.
\newblock In {\em Proceedings of the Student Research Workshop at the 13th
  Conference of the European Chapter of the Association for Computational
  Linguistics\/}. Association for Computational Linguistics, pages 22--31.

\bibitem[{Peng et~al.(2017{\natexlab{a}})Peng, Li, Gao, Liu, Chen, and
  Wong}]{peng2017adversarial}
Baolin Peng, Xiujun Li, Jianfeng Gao, Jingjing Liu, Yun-Nung Chen, and Kam-Fai
  Wong. 2017{\natexlab{a}}.
\newblock Adversarial advantage actor-critic model for task-completion dialogue
  policy learning.
\newblock {\em arXiv preprint arXiv:1710.11277\/} .

\bibitem[{Peng et~al.(2017{\natexlab{b}})Peng, Li, Li, Gao, Celikyilmaz, Lee,
  and Wong}]{peng2017composite}
Baolin Peng, Xiujun Li, Lihong Li, Jianfeng Gao, Asli Celikyilmaz, Sungjin Lee,
  and Kam-Fai Wong. 2017{\natexlab{b}}.
\newblock Composite task-completion dialogue policy learning via hierarchical
  deep reinforcement learning.
\newblock In {\em Proceedings of the 2017 Conference on Empirical Methods in
  Natural Language Processing\/}. pages 2221--2230.

\bibitem[{Peng and Williams(1993)}]{peng1993efficient}
Jing Peng and Ronald~J Williams. 1993.
\newblock Efficient learning and planning within the dyna framework.
\newblock {\em Adaptive Behavior\/} 1(4):437--454.

\bibitem[{Pietquin et~al.(2011)Pietquin, Geist, Chandramohan
  et~al.}]{pietquin2011sample}
Olivier Pietquin, Matthieu Geist, Senthilkumar Chandramohan, et~al. 2011.
\newblock Sample efficient on-line learning of optimal dialogue policies with
  kalman temporal differences.
\newblock In {\em IJCAI Proceedings-International Joint Conference on
  Artificial Intelligence\/}. volume~22, page 1878.

\bibitem[{Pietquin and Hastie(2013)}]{pietquin2013survey}
Olivier Pietquin and Helen Hastie. 2013.
\newblock A survey on metrics for the evaluation of user simulations.
\newblock {\em The knowledge engineering review\/} .

\bibitem[{Racani{\`e}re et~al.(2017)Racani{\`e}re, Weber, Reichert, Buesing,
  Guez, Rezende, Badia, Vinyals, Heess, Li et~al.}]{racaniere2017imagination}
S{\'e}bastien Racani{\`e}re, Th{\'e}ophane Weber, David Reichert, Lars Buesing,
  Arthur Guez, Danilo~Jimenez Rezende, Adri{\`a}~Puigdom{\`e}nech Badia, Oriol
  Vinyals, Nicolas Heess, Yujia Li, et~al. 2017.
\newblock Imagination-augmented agents for deep reinforcement learning.
\newblock In {\em Advances in Neural Information Processing Systems\/}. pages
  5694--5705.

\bibitem[{Schatzmann et~al.(2007)Schatzmann, Thomson, Weilhammer, Ye, and
  Young}]{schatzmann2007agenda}
Jost Schatzmann, Blaise Thomson, Karl Weilhammer, Hui Ye, and Steve Young.
  2007.
\newblock Agenda-based user simulation for bootstrapping a pomdp dialogue
  system.
\newblock In {\em NAACL 2007; Companion Volume, Short Papers\/}. Association
  for Computational Linguistics, pages 149--152.

\bibitem[{Silver et~al.(2016{\natexlab{a}})Silver, Huang, Maddison, Guez,
  Sifre, Van Den~Driessche, Schrittwieser, Antonoglou, Panneershelvam, Lanctot
  et~al.}]{silver2016mastering}
David Silver, Aja Huang, Chris~J Maddison, Arthur Guez, Laurent Sifre, George
  Van Den~Driessche, Julian Schrittwieser, Ioannis Antonoglou, Veda
  Panneershelvam, Marc Lanctot, et~al. 2016{\natexlab{a}}.
\newblock Mastering the game of go with deep neural networks and tree search.
\newblock {\em Nature\/} 529(7587):484--489.

\bibitem[{Silver et~al.(2017)Silver, Schrittwieser, Simonyan, Antonoglou,
  Huang, Guez, Hubert, Baker, Lai, Bolton et~al.}]{silver2017mastering}
David Silver, Julian Schrittwieser, Karen Simonyan, Ioannis Antonoglou, Aja
  Huang, Arthur Guez, Thomas Hubert, Lucas Baker, Matthew Lai, Adrian Bolton,
  et~al. 2017.
\newblock Mastering the game of go without human knowledge.
\newblock {\em Nature\/} 550(7676):354.

\bibitem[{Silver et~al.(2016{\natexlab{b}})Silver, van Hasselt, Hessel, Schaul,
  Guez, Harley, Dulac-Arnold, Reichert, Rabinowitz, Barreto
  et~al.}]{silver2016predictron}
David Silver, Hado van Hasselt, Matteo Hessel, Tom Schaul, Arthur Guez, Tim
  Harley, Gabriel Dulac-Arnold, David Reichert, Neil Rabinowitz, Andre Barreto,
  et~al. 2016{\natexlab{b}}.
\newblock The predictron: End-to-end learning and planning.
\newblock {\em arXiv preprint arXiv:1612.08810\/} .

\bibitem[{Singh et~al.(2002)Singh, Litman, Kearns, and
  Walker}]{singh2002optimizing}
Satinder Singh, Diane Litman, Michael Kearns, and Marilyn Walker. 2002.
\newblock Optimizing dialogue management with reinforcement learning:
  Experiments with the njfun system.
\newblock {\em Journal of Artificial Intelligence Research\/} 16:105--133.

\bibitem[{Singh(1992)}]{singh1992reinforcement}
Satinder~P Singh. 1992.
\newblock Reinforcement learning with a hierarchy of abstract models.
\newblock In {\em Proceedings of the National Conference on Artificial
  Intelligence\/}. JOHN WILEY \& SONS LTD, 10, page 202.

\bibitem[{Su et~al.(2016{\natexlab{a}})Su, Gasic, Mrksic, Rojas-Barahona,
  Ultes, Vandyke, Wen, and Young}]{su2016continuously}
Pei-Hao Su, Milica Gasic, Nikola Mrksic, Lina Rojas-Barahona, Stefan Ultes,
  David Vandyke, Tsung-Hsien Wen, and Steve Young. 2016{\natexlab{a}}.
\newblock Continuously learning neural dialogue management.
\newblock {\em arXiv preprint arXiv:1606.02689\/} .

\bibitem[{Su et~al.(2016{\natexlab{b}})Su, Gasic, Mrksic, Rojas-Barahona,
  Ultes, Vandyke, Wen, and Young}]{su2016line}
Pei-Hao Su, Milica Gasic, Nikola Mrksic, Lina Rojas-Barahona, Stefan Ultes,
  David Vandyke, Tsung-Hsien Wen, and Steve Young. 2016{\natexlab{b}}.
\newblock On-line active reward learning for policy optimisation in spoken
  dialogue systems.
\newblock {\em arXiv preprint arXiv:1605.07669\/} .

\bibitem[{Sutton(1990)}]{sutton1990integrated}
Richard~S Sutton. 1990.
\newblock Integrated architectures for learning, planning, and reacting based
  on approximating dynamic programming.
\newblock In {\em Proceedings of the seventh international conference on
  machine learning\/}. pages 216--224.

\bibitem[{Sutton and Barto(1998)}]{sutton1998introduction}
Richard~S Sutton and Andrew~G Barto. 1998.
\newblock {\em Introduction to reinforcement learning\/}, volume 135.
\newblock MIT press Cambridge.

\bibitem[{Sutton et~al.(2012)Sutton, Szepesv{\'a}ri, Geramifard, and
  Bowling}]{sutton2012dyna}
Richard~S Sutton, Csaba Szepesv{\'a}ri, Alborz Geramifard, and Michael~P
  Bowling. 2012.
\newblock Dyna-style planning with linear function approximation and
  prioritized sweeping.
\newblock {\em arXiv preprint arXiv:1206.3285\/} .

\bibitem[{Tamar et~al.(2016)Tamar, Wu, Thomas, Levine, and
  Abbeel}]{tamar2016value}
Aviv Tamar, Yi~Wu, Garrett Thomas, Sergey Levine, and Pieter Abbeel. 2016.
\newblock Value iteration networks.
\newblock In {\em Advances in Neural Information Processing Systems\/}. pages
  2154--2162.

\bibitem[{Wen et~al.(2015)Wen, Gasic, Mrksic, Su, Vandyke, and
  Young}]{wen2015semantically}
Tsung-Hsien Wen, Milica Gasic, Nikola Mrksic, Pei-Hao Su, David Vandyke, and
  Steve Young. 2015.
\newblock Semantically conditioned lstm-based natural language generation for
  spoken dialogue systems.
\newblock {\em arXiv preprint arXiv:1508.01745\/} .

\bibitem[{Williams et~al.(2017)Williams, Asadi, and Zweig}]{williams2017hybrid}
Jason~D Williams, Kavosh Asadi, and Geoffrey Zweig. 2017.
\newblock Hybrid code networks: Practical and efficient end-to-end dialog
  control with supervised and reinforcement learning.
\newblock In {\em Proceedings of the 55th Annual Meeting of the Association for
  Computational Linguistics\/}.

\bibitem[{Young et~al.(2013)Young, Ga{\v{s}}i{\'c}, Thomson, and
  Williams}]{young2013pomdp}
Steve Young, Milica Ga{\v{s}}i{\'c}, Blaise Thomson, and Jason~D Williams.
  2013.
\newblock Pomdp-based statistical spoken dialog systems: A review.
\newblock {\em Proceedings of the IEEE\/} 101(5):1160--1179.

\bibitem[{Zhao and Eskenazi(2016)}]{zhao2016towards}
Tiancheng Zhao and Maxine Eskenazi. 2016.
\newblock Towards end-to-end learning for dialog state tracking and management
  using deep reinforcement learning.
\newblock {\em arXiv preprint arXiv:1606.02560\/} .

\end{thebibliography}
\bibliographystyle{acl_natbib}

\newpage
\appendix

\section{Dataset Annotation Schema}
\label{app:data_schema}
Table~\ref{tab:data_schema} lists all annotated dialogue acts and slots in details.

\begin{table}[h]
\small
\begin{tabular}{|c|l|l|}
\hline
\multicolumn{2}{|c|}{Annotations} \\
\hline\hline
& \textsf{request}, \textsf{inform}, \textsf{deny}, \textsf{confirm\_question},\\ 
Intent & \textsf{confirm\_answer}, \textsf{greeting}, \textsf{closing}, \textsf{not\_sure},\\
 & \textsf{multiple\_choice}, \textsf{thanks}, \textsf{welcome} \\
\hline
\multirow{4}{*}{Slot} & \textsf{city}, \textsf{closing}, \textsf{date}, \textsf{distanceconstraints}, \\
& \textsf{greeting}, \textsf{moviename}, \textsf{numberofpeople}, \\ 
& \textsf{price}, \textsf{starttime}, \textsf{state}, \textsf{taskcomplete}, \textsf{theater}, \\
& \textsf{theater\_chain}, \textsf{ticket}, \textsf{video\_format}, \textsf{zip} \\
\hline
\end{tabular}
\centering
\caption{The data annotation schema}
\label{tab:data_schema}
\end{table}

\section{User Simulator}
\label{app:user_sim}
In the task-completion dialogue setting, the entire conversation is around a user goal implicitly, but the agent knows nothing about the user goal explicitly and its objective is to help the user to accomplish this goal. Generally, the definition of user goal contains two parts: 

\begin{compactitem}
\item \emph{inform\_slots} contain a number of slot-value pairs which serve as constraints from the user.
\item \emph{request\_slots} contain a set of slots that user has no information about the values, but wants to get the values from the agent during the conversation. \textsf{ticket} is a default slot which always appears in the \emph{request\_slots} part of user goal.
\end{compactitem}

To make the user goal more realistic, we add some constraints in the user goal: slots are split into two groups. Some of slots must appear in the user goal, we called these elements as \emph{Required slots}. In the movie-booking scenario, it includes \textsf{moviename, theater, starttime, date, numberofpeople}; the rest slots are \emph{Optional slots}, for example, \textsf{theater\_chain, video\_format} etc.

We generated the user goals from the labeled dataset mentioned in Section~\ref{sec:dataset}, using two mechanisms. One mechanism is to extract all the slots (known and unknown) from the first user turns (excluding the greeting user turn) in the data, since usually the first turn contains some or all the required information from user. The other mechanism is to extract all the slots (known and unknown) that first appear in all the user turns, and then aggregate them into one user goal. We dump these user goals into a file as the user-goal database. Every time when running a dialogue, we randomly sample one user goal from this user goal database.

\end{document}